\pdfoutput=1

\documentclass[11pt]{article}

\usepackage[final]{acl}

\usepackage{times}
\usepackage{latexsym}
\usepackage{amsmath}
\usepackage{subcaption}
\usepackage[T1]{fontenc}

\usepackage[utf8]{inputenc}

\usepackage{microtype}

\usepackage{inconsolata}

\usepackage{graphicx}
\usepackage{adjustbox}
\usepackage{siunitx}
\usepackage{booktabs}

\usepackage{placeins}

\def\InfR{{InfiR}}

\title{\InfR~: Crafting Effective Small Language Models and Multimodal Small Language Models in Reasoning}

\author{
Congkai Xie\textsuperscript{1},
Shuo Cai\textsuperscript{4},
Wenjun Wang\textsuperscript{4},
Pengxiang Li\textsuperscript{6},
Zhijie Sang\textsuperscript{1},
Kejing Yang\textsuperscript{1},
Yiming Zhang\textsuperscript{2},\\
\textbf{
Zhen Li\textsuperscript{1,2},
Guanghao Zhu\textsuperscript{7},
Zeyu Liu\textsuperscript{5},
Yang Yu\textsuperscript{8},
Yuhang Liu\textsuperscript{3},
Su Lu\textsuperscript{2},
Baoyi He\textsuperscript{3},
Qi Zhou\textsuperscript{5},
}\\
\textbf{
Xiaotian Han\textsuperscript{9},
Jianbo Yuan\textsuperscript{10},
Shengyu Zhang\textsuperscript{3},
Fei Wu\textsuperscript{3},
Hongxia Yang\textsuperscript{1,2}
}\\
\textsuperscript{1} Reallm Labs,
\textsuperscript{2} The Hong Kong Polytechnic University,
\textsuperscript{3} Zhejiang University,\\
\textsuperscript{4} South China University of Technology,
\textsuperscript{5} Harbin Institute of Technology,\\
\textsuperscript{6} Dalian University of Technology,
\textsuperscript{7} University of Electronic Science and Technology of China,\\
\textsuperscript{8} Beijing University of Posts and Telecommunications,
\textsuperscript{9} TikTok,
\textsuperscript{10} Amazon\\
\textsuperscript{*}\,Corresponding authors. \quad
\texttt{hongxia.yang@polyu.edu.hk}
}
\vspace{0.2in}

\begin{document}
\maketitle
\begin{abstract}
Large Language Models (LLMs) and Multimodal Large Language Models (MLLMs) have made significant advancements in reasoning capabilities. However, they still face challenges such as high computational demands and privacy concerns. This paper focuses on developing efficient Small Language Models (SLMs) and Multimodal Small
Language Models (MSLMs) that retain competitive reasoning abilities. We introduce a novel training pipeline that enhances reasoning capabilities and facilitates deployment on edge devices, achieving state-of-the-art performance while minimizing development costs. \InfR~ aims to advance AI systems by improving reasoning, reducing adoption barriers, and addressing privacy concerns through smaller model sizes. Resources are available at https://github.
com/Reallm-Labs/InfiR.
\end{abstract}

\section{Introduction}
\vspace{-0.1in}
\noindent In recent years, Large Language Models (LLMs) ~\cite{touvron2023llamaopenefficientfoundation, dsvi,dsvii,qwen,qwen2,qwen2025qwen25technicalreport} and Multimodal Large Language Models (MLLMs)~\cite{qwen2vl2024,internvl25} have shown remarkable advancements in reasoning capabilities. However, these models, often consisting of hundreds of billions of parameters, pose significant challenges related to computational resources, deployment costs, and environmental impact. Their training and development demand substantial infrastructure investments, making them accessible primarily to major technology corporations. Additionally, these models usually require cloud deployment, which raises privacy concerns regarding user data.

\noindent Small models, typically comprising fewer than 2 billion parameters, have emerged as a promising solution to these challenges. These models strive to achieve an optimal balance between performance and efficiency. Their considerably lower training and development costs increase accessibility for researchers and organizations. There is a key challenge for SLMs and MSLMs in enhancing reasoning capabilities while reducing the size of the model.

\noindent This paper focuses on: (1) developing efficient small language models with competitive reasoning capabilities and (2) extending these models to handle multimodal inputs while managing real-world operational system tasks. We explore novel training pipelines that enable these compact models to perform reasoning tasks while being deployable on edge devices. Our contributions to the field include the following:
\vspace{-0.1in}
\begin{itemize}
\item We propose a pre- and post-training pipeline for small models that enhances reasoning capabilities, completing training in under 6000 GPU hours.
\item Our \InfR-1B-Base and \InfR-1B-Instruct models achieve state-of-the-art performance at the 1B parameter scale, with reasoning-related average score improvements of 2.26x and 1.33x over Llama3.2-1B-Base and Llama3.2-1B-Instruct.
\item Our \InfR-VL-1.6B model achieves the state-of-the-art performance in the Android World scenario, with an accuracy of 28\% increment compared to the best SOTA among small models.
\end{itemize}
\vspace{-0.1in}
\noindent Through this research, we aim to advance the development of practical, efficient AI systems capable of performing reasoning tasks, while reducing barriers to AI system adoption and addressing user privacy requirements through model size reduction.

\begin{figure*}[!h]
  \centering
      \includegraphics[width=\linewidth]
      {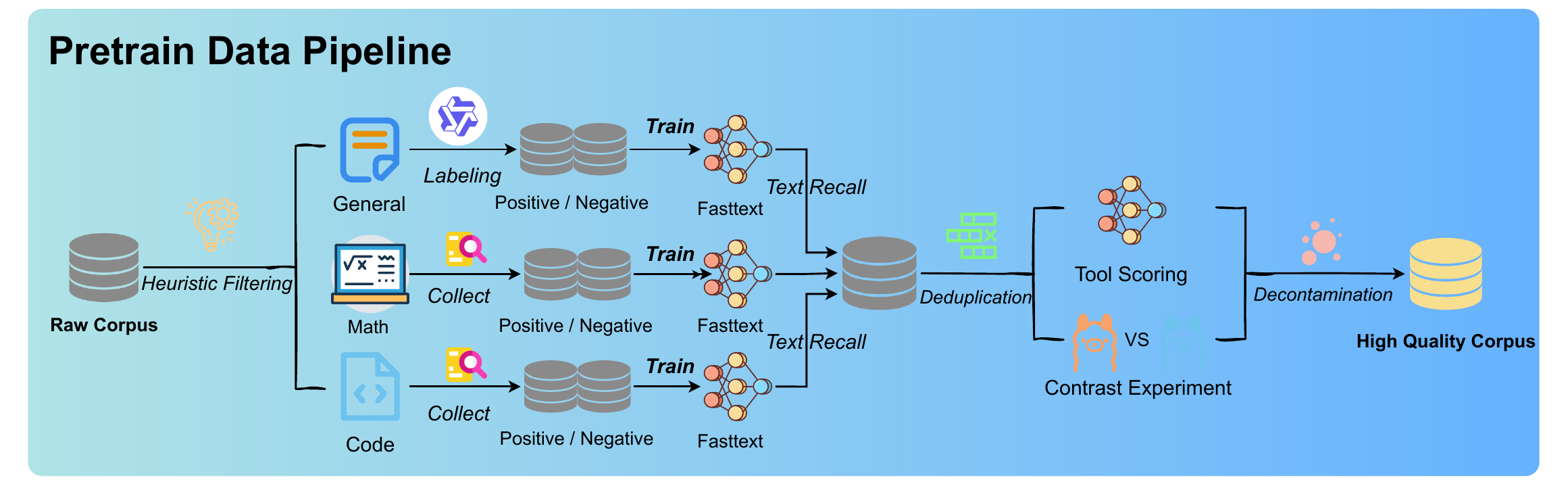}
      \caption{The pipeline of pretrain data drocesses: heuristic filtering, reasoning-oriented text recall, deduplication, quality assessment and decontamination. Comparative experiments on LLaMA3.2-1B with differently cleaned datasets validate the significance of data quality.}
    \label{fig:pretrain_data_pipeline}
\end{figure*}
\vspace{-0.1in}
\section{Small Language Model Pre-training}
\vspace{-0.1in}
\subsection{Pre-training Data}
\vspace{-0.1in}
\noindent To construct effective SLMs with strong reasoning capabilities, high-quality pre-training data are essential. We present a comprehensive pipeline for collecting and processing such data.

\subsubsection{Data Collection}
\noindent To develop SLMs with enhanced reasoning capabilities, we constructed a comprehensive dataset comprising two primary components: source code data and reasoning-oriented text data. This approach is designed to capture both programmatic logic and natural language reasoning patterns. For the source code component, we curated the dataset by combining several high-quality open-source resources, including The Stack v2 \cite{lozhkov2024starcoder} and Starcoder \cite{li2023starcoder} and additional Python-centric repositories from GitHub. More details of the distribution of programming languages can be found in Figure \ref{fig:pl_dis}. To acquire reasoning-oriented text data, we assembled a diverse corpus encompassing web content, academic literature, books, and encyclopedic sources. This corpus was specifically curated to emphasize content that shows logical reasoning, analytical thinking, and structured argumentation. 
\vspace{-0.1in}
\subsubsection{Data Pipeline}
\vspace{-0.1in}
\noindent From a technical perspective, as the model size decreases, its information storage capacity diminishes proportionally. To develop a SLM specifically targeted for reasoning tasks, it is essential to construct a sophisticated data pipeline that can effectively filter out noise while preserving reasoning-related information. The workflow of the entire pipeline architecture is depicted in Figure \ref{fig:pretrain_data_pipeline}.

\noindent \textbf{Heuristic filtering}
In the initial phase, we employ heuristic filters to perform preliminary filtering on the raw corpus, thereby reducing noise in the data. We employed heuristic filters from FineWeb~\cite{penedo2024finewebdatasetsdecantingweb} to extract high-quality text from web pages. For code data, we selectively filtered and processed repositories written in widely used languages such as Python, JavaScript, Java, and C to ensure relevance and quality. Rule-based filters~\cite{opencoder24} were applied to remove files that were contaminated or significantly deviated from standard patterns.


\noindent \textbf{Reasoning-oriented text recall} 
\noindent Reasoning-oriented data, unlike code, lack explicit patterns that can be identified through rule-based approaches. Instead, model-based methods are typically required to extract potential reasoning relationships from large volumes of text. Our reasoning-oriented data retrieval strategy is categorized into three components. First, we prioritize the retrieval of mathematics-related samples, as mathematical texts often exhibit strong reasoning characteristics. Second, we recognize that a significant portion of code-related content in internet data contains reasoning elements. Finally, we aim to retrieve reasoning-related texts from various other domains.

\noindent We developed a standard text recall data pipeline, drawing inspiration from DeepSeekmath~\cite{shao2024deepseekmathpushinglimitsmathematical} and OpenCoder~\cite{huang2024opencoderopencookbooktoptier}. The pipeline begins with establishing seed datasets. For mathematics-related content, we utilize high-quality mathematical datasets such as OpenWebMath~\cite{paster2023openwebmath} and InfiMM-WebMath~\cite{han2024infimm} as seed data. For code-related text, we employ StackOverflow as the seed dataset. For other domains, we utilize Qwen2.5-7B-Instruct~\cite{qwen2025qwen25technicalreport} to annotate text URLs and titles, while also incorporating responses synthesized by large language models from datasets like Infinity Instruct as seed data.

\noindent Following seed data acquisition, we train domain-specific fasttext models using positive samples from the seed data and random negative samples from web pages and books. These fasttext models are then used to recall relevant content from the remaining corpus.

\noindent \textbf{Deduplication} An excessive amount of homogeneous data can be detrimental to small language models. therefore, it is crucial to ensure data diversity. 
\noindent To optimize data efficiency while maintaining semantic diversity, we implemented a global MinHash algorithm to efficiently detect and eliminate near-duplicate documents.

\noindent \textbf{Quality assessment} To ensure comprehensive data quality across diverse domains, we establish a two-step evaluation framework. 
First, we employ a domain-specific quality assessment tool: (1) For web content, we leverage FineWeb-edu-scorer~\cite{penedo2024finewebdatasetsdecantingweb} to evaluate document quality through multiple dimensions; (2) For mathematical content, we utilize a model-based classifier following~\cite{lozhkov2024finemath} that scores reasoning and deduction capabilities on a 1-5 scale, filtering to retain only high-quality samples; (3) For code data quality validation, we utilize a static analysis tool to validate syntactic correctness and identify potential structural issues.
\noindent We also perform comprehensive ablation experiments through continued pre-training on LLaMA 3.2-1B, comparing the performance between models trained on raw and filtered datasets.
The model trained on filtered data exhibits better performance across multiple code-related benchmarks compared to its counterpart trained on the raw dataset. 
\noindent Moreover, the filtering process facilitates accelerated convergence, enabling the model to achieve desired performance metrics with fewer training steps.


\noindent \textbf{Decontamination} To ensure fairness of comparison, we implemented a token-level 10-gram decontamination algorithm to remove potentially contaminated content in standard benchmarks.
\vspace{-0.1in}
\subsection{Annealing Data}
\vspace{-0.1in}
\noindent After training on 900 billion tokens, we aimed to further enhance reasoning abilities and bridge the gap between the pretraining and supervised fine-tuning stages. To achieve this, we constructed an annealing dataset comprising additional high-quality reasoning data and continued training the checkpoint from the previous stage on 40 billion annealing data to obtain the final base model.

\noindent \textbf{Original data} During the annealing phase, we maintained the original proportion of source code. For reasoning-oriented text, we removed most web page data, retaining only a small portion related to mathematics and code.

\noindent \textbf{Open source annealing data} We collected annealing data used by Dolmino~\cite{olmo20242olmo2furious} and OpenCoder. We also include a high-quality code-related datasets, incorporating training samples from APPS ~\cite{hendrycks2021measuringcodingchallengecompetence} and Code Contest~\cite{Li_2022}.
For each programming problem across the above datasets, we retain a single solution and prioritize Python implementations. These high-quality datasets were utilized in comparative experiments, which demonstrated a significant improvement in the base model's performance on reasoning benchmarks in few-shot settings.

\noindent \textbf{Synthetic data} We further leveraged a large language model to generate a batch of synthetic data, implementing stringent selection mechanisms to ensure high quality. For reasoning-oriented data, we employed a reward model with rejection sampling to enhance logical coherence and correctness. For code-related data, we validated its syntax and functionality within a sandbox environment, ensuring robustness and reliability.

\vspace{-0.1in}
\subsection{Offline Evaluations for Data-mixing}
\vspace{-0.1in}
\subsubsection{Pre-training Stage}
\noindent 
During the pre-training phase, the model compresses knowledge into its parameters. We prioritize the model's \textbf{recall} capability, primarily using Negative Log Likelihood (NLL) to assess performance. NLL evaluations are conducted on two types of data:

\noindent \textbf{Sample from webpage} During pretraining, we utilize a large amount of web page data. To measure model convergence, we sample a validation set from the web data. Although the web data has been cleaned, there might still be meaningless text in the samples. Therefore, we use existing large language models to calculate the NLL of the text and rank them separately. When a text ranks in the bottom 20\% across all rankings, we consider it meaningless and discard it. The NLL for both types is defined as:
\begin{eqnarray}
 \mbox{NLL} = -\sum_{i=1}^{n} \log P(t_i|t_{<i}),   
\end{eqnarray}
where $t_i$ represents the i-th token in the webpage text sequence, and $P(t_i|t_{<i})$ is the model's predicted probability for that token given the previous tokens.

\noindent \textbf{Downstream Task Benchmark} We utilize the NLL per token on the correct answers from downstream benchmarks as our metric across tasks. For multiple-choice tasks, we normalize the probabilities assigned to all options to adjust the NLL, creating a metric that both correlates with accuracy improvements and shows stable improvement as model scale increases. the normalized probability is:
\begin{eqnarray}
P_{\mbox{normalized}}(c_i) = \frac{e^{s_i}}{\sum_{j=1}^{k} e^{s_j}},    
\end{eqnarray} 
where $s_i$ is the model's score for choice $i$, and $k$ is the number of choices.
\begin{eqnarray}
    ppl_p(t_{1:n}) = \exp(\frac{1}{n}\sum_{i=1}^n\mathrm{log}\frac{1}{p(t_i | t_{1:i-1})})
\end{eqnarray}
\subsubsection{Annealing Stage}
\noindent During the annealing stage, the model rapidly converges due to the decay of the learning rate. \textbf{Precision} is prioritized over \textbf{recall}. To evaluate the model's performance, we use a downstream benchmark for few-shot generation instead of the NLL method.

\noindent We can suppose a contaminated distribution \( r \) is introduced into the original clean distribution \( p \) with a probability of \( \epsilon \), leading to a new distribution \( q \), which satisfies:

{\small{ 
\begin{multline}
q(t_i|t_{1:i-1}) = (1-\epsilon) p(t_i|t_{1:i-1})   + \epsilon r(t_i|t_{1:i-1}).
\end{multline}}}

\noindent By computing the $ppl$ for the new distribution $q$, than we have:


\begin{align}
ppl_q(t_{1:n}) & =
\exp(\frac{1}{n}\sum_{i=1}^n\mathrm{log}\frac{1}{q(t_i|t_{1:i-1})}) \nonumber\\
\end{align}
\noindent Since $q(t_i|t_{1:i-1}) \geq (1-\epsilon)p(t_i |t_{1:i-1})$, we can obtain
\begin{align}
    ppl_q(t_{1:n}) & \leq \exp(\frac{1}{n}\sum_{i=1}^n\mathrm{log}\frac{1}{(1-\epsilon)  p(t_i|t_{1:i-1})}) \nonumber \\
    & \leq \frac{1}{(1-\epsilon)}ppl_p(t_{1:n}) \nonumber \\ 
    & \approx (1+\epsilon)ppl_p(t_{1:n}).
\end{align}
\noindent The approximation holds when $\epsilon$ is small enough. This indicates that if 5\% contaminated distribution is introduced, the perplexity increases no more than 5\%. However, this could lead to very poor results in generation for the language model, since
the model may produce a corrupted token approximately every 20 tokens on average. 
\noindent Therefore, during the annealing stage, we evaluate the model directly using a downstream benchmark in a few-shot setting to determine a data mixing ratio that optimally balances various capabilities.

\vspace{-0.1in}
\subsection{Training Details}
\vspace{-0.1in}
\noindent Ultimately, we curated a high-quality reasoning-related dataset consisting of approximately 900 billion tokens for pretraining and 40 billion tokens for annealing. Details of the dataset's composition are provided in the Table~\ref{tab:pretraining_data_composition} and Table~\ref{tab:annealing_data_composition}. Based on LLaMA-3.2 1B architecture, we trained our model in two phases. During the pretraining phase, we trained the model on the curated 900 billion tokens for one epoch, followed by an annealing phase on an additional 40 billion tokens for one epoch. We employed a learning rate of 1.4e-3 with a size of 2048 and a sequence length of 4096 tokens throughout the training process. Training was conducted using NVIDIA NeMo~\cite{kuchaiev2019nemotoolkitbuildingai} with distributed optimization and DDP gradient overlap on a cluster of 64 H800 GPUs over a total of 90 hours, equating to 5760 GPU hours.

\begin{figure*}[!h]
  \centering
      \includegraphics[width=\linewidth]
      {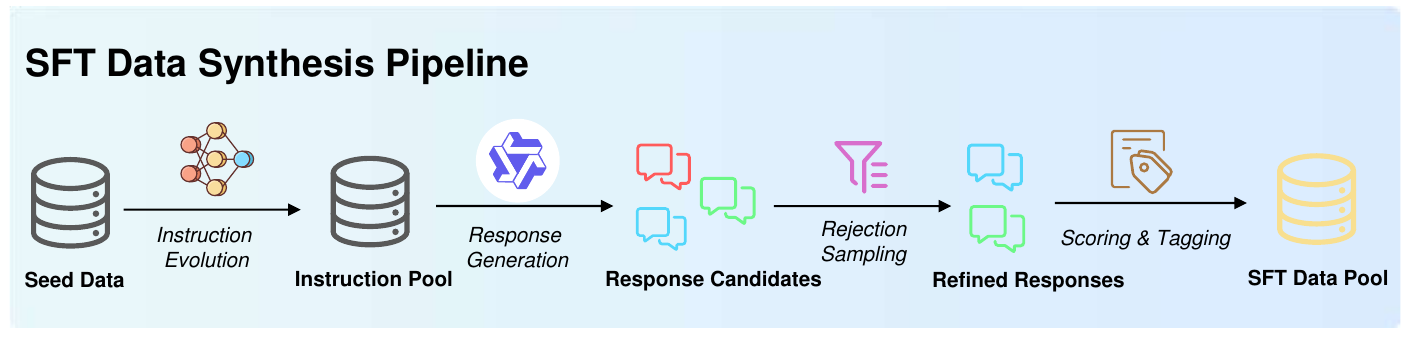}
      \caption{Supervised fine-tuning data synthesis pipeline. The pipeline initiates with a set of high-quality seed data, which is augmented through instruction evolution. Response candidates are generated using the Qwen-2.5-32B-Instruct model, followed by rejection sampling with a reward model and sandbox environment. Finally, we score the curated data for quality and difficulty, and assign domain labels.}
        \label{fig:sft_data_pipeline}
\end{figure*}

\vspace{-0.1in}
\section{Small Language Model Post-training}
\vspace{-0.1in}
\noindent During the post-training phase of the language model, we leverage Supervised Fine-Tuning (SFT) with meticulously curated datasets to enhance instruction-following and reasoning capabilities. The deliberate construction of SFT data, which accounts for the balance across diverse domains, facilitates enhancements in controllability and robustness, ensuring optimal performance in various applications. 
\subsection{ Supervised Fine-tuning Data}
\vspace{-0.1in}
\noindent The quality and diversity of data play a crucial role in supervised fine-tuning. We utilize high-quality, publicly available datasets involving instruction-following, reasoning, and code-related tasks, including Infinity-Instruct~\cite{InfinityInstruct2024}, Orca-AgentInstruct-1M-v1~\cite{agentinstruct}, ScaleQuest~\cite{ding2024unleashing}, NuminaMath~\cite{li2024numinamath}, and OPC-SFT-Stage2~\cite{opencoder24}. In addition, we developed a data synthesis pipeline to generate high-quality SFT data, incorporating components such as instruction evolution, rejection sampling, etc., as illustrated in Figure~\ref{fig:sft_data_pipeline}.

\noindent We curated a high-quality set of instructions from publicly available datasets as seed data and employed a large language model for instruction evolution, generating more diverse and complex instructions. Subsequently, we utilized the Qwen-2.5-32B-Instruct model to generate responses corresponding to these instructions. We encouraged the model to engage in a "step-by-step" reasoning process prior to delivering the final answer, which promoted a more rigorous and logically consistent output.

\noindent \textbf{Rejection sampling} For each instruction, we generate multiple responses and employ rejection sampling to ensure the quality of the data. For logic and mathematical reasoning data, we utilize a reward model to assess  responses and select the one with the highest score. For code data, we perform execution-based code verification within a sandbox environment to ensure the correctness and functionality of the generated code.

\noindent \textbf{Diversity and quality} We assign domain labels to each training sample and perform diversity sampling from the extensive dataset to balance the distribution of data across different domains. Using heuristic rules and LLM-based scoring judgments, we filter out low-quality instructions and responses. Additionally, we use a difficulty scoring model to assess mathematical reasoning data, maintaining a balanced distribution across varying levels of difficulty. Through the combination of diverse and high-quality data sources, instruction augmentation, rejection sampling, and rigorous quality control, we ensure that the training data is both comprehensive and reliable. These carefully designed methodologies contribute to enhancing the model's reasoning capabilities, enabling it to better handle complex tasks and exhibit improved performance across a wide range of problem domains.

\vspace{-0.1in}
\subsection{Training Details}
\vspace{-0.1in}
\noindent We employed the Llama 3 chat template and applied supervised fine-tuning using a standard cross-entropy loss on a few million samples, while masking the loss on the prompt tokens. We fine-tuned the model for 4 epochs with a learning rate of 2e-5 and a batch size of 128. We utilized the cosine learning rate scheduler with a warm-up ratio of 0.1 to optimize training process. The model was fine-tuned on the Llama 3 template with a maximum sequence length of 4096 tokens.
\vspace{-0.1in}

\vspace{-0.1in}
\section{Small Multimodal Language Model Training}
\vspace{-0.1in}
\subsection{Multimodal Pre-training}
\vspace{-0.1in}



\subsubsection{Data Collection}
\vspace{-0.1in}
\noindent To systematically develop multimodal reasoning capabilities, we implement a hierarchical data collection strategy that progressively builds from fundamental visual-language alignment to complex reasoning tasks. At the foundation level, we collect data from 11 task domains including captioning, general QA, and OCR to train the MLP projector while keeping the vision encoder and language model frozen, establishing the essential mapping between visual features and LLM's textual semantic space. In addition, we enhance the model's text comprehension abilities by synthetic rendering text data, converting documents, code snippets, and other textual content into visual formats. 

\begin{figure*}[!h]
  \centering
      \includegraphics[width=\linewidth]
      {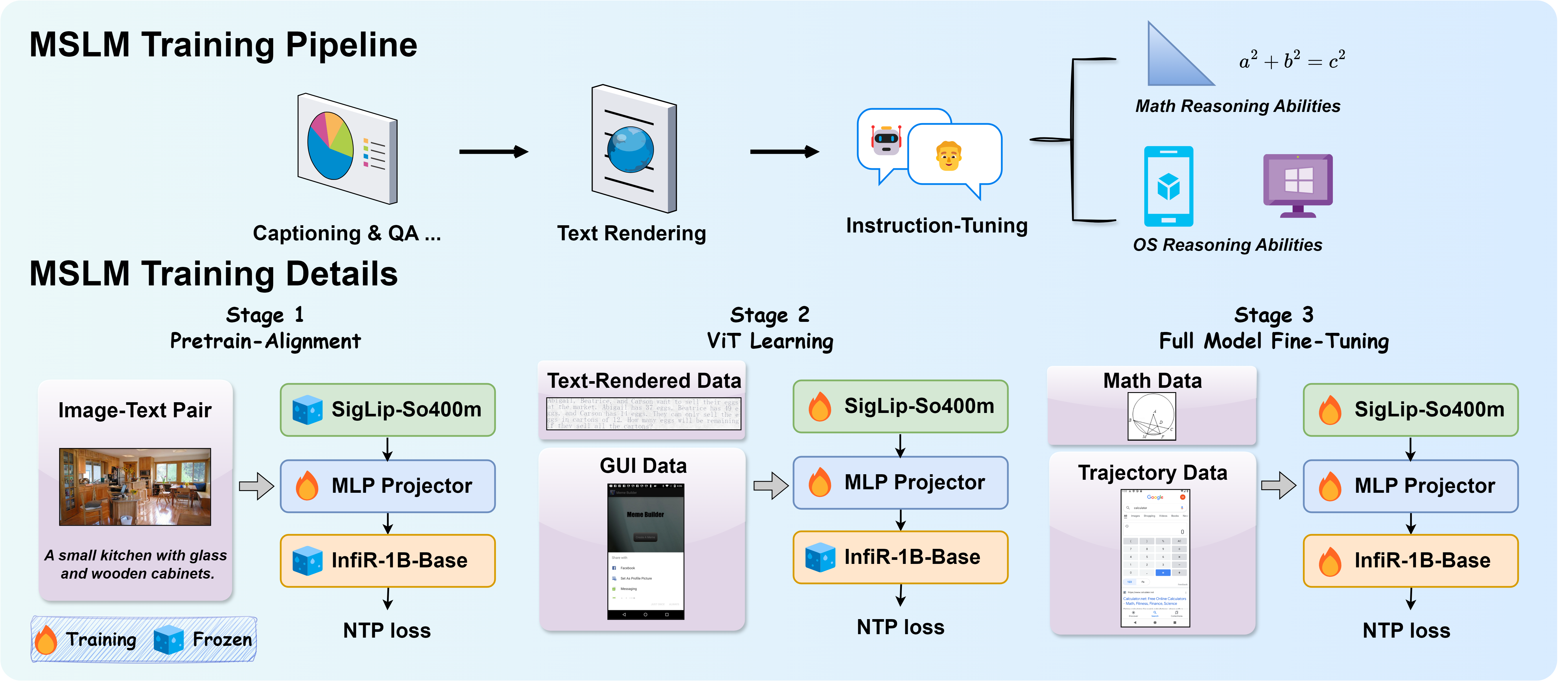}
      \caption{Illustration of the MSLM training pipeline and the MSLM training details, showcasing the progression from captioning and QA tasks to text rendering, followed by instruction-tuning, culminating in enhanced mathematical and operating system reasoning abilities.}
\label{fig:mslm_train}
\vspace{-3mm}
\end{figure*}
\vspace{-0.1in}
\subsubsection{Data Cleaning}
\vspace{-0.1in}
\noindent In order to reduce the impact of low-quality, low-similarity image-text pairs in the dataset on the model's ability to understand the semantics of images and texts, we use the Vista model \cite{vista} to extract the embeddings of the data pairs, calculate the similarity by pair, and then select a reasonable similarity threshold to filter out low-quality, low-similarity image-text pairs. The details is show in Appendix~\ref{appendix:mulimodal_data_cleaning}.
\vspace{-0.1in}
\subsection{Multimodal Instruction-Tuning}
\vspace{-0.1in}
\subsubsection{General Vision Reasoning Abilities}
\vspace{-0.1in}
As we did in the SFT of SLMs, during the SFT stage of MSLMs, we collected multiple open-source instruction-tuning datasets spanning 14 domains, including general QA, mathematics, charts, OCR, documents, science, medical, GUI, code, etc., while covering multiple modalities such as single-image, multi-image, and text, as shown in Table \ref{tab:mm_rsn_ds}, and synthesized more high-quality vision reasoning data through multiple methods such as text rendering and virtual environment, so as to enable the model to have comprehensive general vision reasoning capabilities for image-related problems in the real world and strengthen the effective semantic alignment between visual input and SLM.
\vspace{-5mm}
\subsubsection{Multimodal Math Reasoning Abilities}
\vspace{-0.1in}
\noindent After the initial stimulation of general vision reasoning ability, we fine-tuned SLM in different subdivisions for different scenarios to enhance the ability of MSLM in the corresponding fields. In mathematics, we used multiple data sets such as InfiMM-WebMath-40B \cite{han2024infimm}, MathVista \cite{lu2024mathvista}, and MM-Math \cite{sun2024mmmath} as bases, and used ChatGPT to synthesize multimodal math question-answering CoT-PoT data; in addition, we also used the text rendering mechanism to mix in some pure text math question-answering Instruction tuning data.

\vspace{-0.1in}
\subsubsection{Operator-System Reasoning Abilities}
\vspace{-0.1in}
\noindent Our MSLM's operator-system reasoning capabilities are developed through a two-stage supervised fine-tuning framework that progressively builds from fundamental understanding to advanced reasoning.~\cite{liu2025infiguiagent} The foundation stage establishes essential GUI comprehension by utilizing diverse datasets spanning GUI understanding, grounding, and question-answering tasks. We standardize coordinate systems to a [0,1000] scale and implement a reference-augmented annotation format, enabling precise spatial reasoning while maintaining natural language flow in the model's responses.
\noindent Building upon this foundation, with existing trajectory data, we synthesize 45K training samples that incorporate these advanced reasoning patterns, enabling our MSLM to handle complex GUI interactions through structured reasoning rather than mere pattern matching.
\vspace{-0.1in}
\subsection{Training Details}
\vspace{-0.1in}
\noindent As shown in Fig \ref{fig:mslm_train}, our proposed Multimodal Small Language Model Architecture integrates a vision encoder with a language model. For the vision encoder, we utilize SigLip-So400m~\cite{zhai2023sigmoid}, which is built upon the SoViT-400M. The language model component is based on our \InfR-1B-Base, which provides reasoning capabilities. To align the visual features with the language model's latent space, we employ a simple MLP layer as the visual projector.

\noindent During the Pretraining stage, we exclusively train the MLP projector with the aim of aligning the backbones of the ViT and LLM. In this phase, we utilize a large dataset of captions.

\noindent In the SFT stage, the training is divided into two sub-stages. The first sub-stage focuses on enhancing the model's capabilities with images. Here, we unfreeze the parameters of the ViT and train both the ViT and the adapter. We employ a substantial amount of text rendering synthetic samples and GUI samples to ensure the model develops initial vision reasoning capabilities. Additionally, we retain some samples from the pretraining phase to maintain the model's general captioning abilities.

\noindent In the second SFT sub-stage, we unfreeze all parameters and train on the most challenging samples, such as trajectory data in operating system scenarios and tool usage samples in mathematical contexts. This step is designed to enhance the model's planning and reasoning abilities.

\begin{table*}[htbp]
\centering
\begin{adjustbox}{width=\linewidth}
\begin{tabular}{lccccccc}
\hline
\textbf{Model} & \textbf{MMLU} & \textbf{GSM8K} & \textbf{MATH} & \textbf{HumanEval} & \textbf{MBPP} & \textbf{MBPP(3-shot)} \\ 
\hline
Llama-3.2-1B & 32.74 (32.2) & 8.11 & 3.42 & 17.68 & 33.46 & 24.8 \\
Qwen-2.5-1.5B & 63.03 (60.9) & 66.57 (68.5) & 31.24 (35.0) & 35.37 (37.2) & 58.37 (60.2) & 41.4 \\
\hline
\InfR-1B-Base & 47.24 & 63.46 & 31.82 & 37.80 & 53.40 & 37.6 \\
\hline
\end{tabular}
\end{adjustbox}
\vspace{-0.1in}
\caption{Performance of base models on various benchmarks using few-shot evaluation. The values in parentheses indicate the claimed results from the respective papers.}
\label{tab:base_performance}
\end{table*}

\begin{table*}[htbp]
\centering
\begin{adjustbox}{width=\linewidth}
\begin{tabular}{lccccccc}
\hline
\textbf{Model} & \textbf{MMLU} & \textbf{GSM8K} & \textbf{MATH} & \textbf{HumanEval} & \textbf{MBPP} \\ 
\hline
Llama-3.2-1B-Instruct & 46.27 (49.3) & 47.9 (44.4) & 30.0 (30.6) & 39.63 & 49.03 \\
Qwen-2.5-1.5B-Instruct & 61.78 & 74.3 (73.2) & 53.4 (55.2) & 51.83 (61.6)  & 56.81 (63.2) \\
\hline
\InfR-1B-Instruct & 50.22 & 70.9 & 46.4 & 58.54 & 56.03 \\
\hline
\end{tabular}
\end{adjustbox}
\vspace{-0.1in}
\caption{Performance of \InfR-1B-Instruct models on various benchmarks using zero-shot evaluation. The values in parentheses indicate the claimed results from the respective papers. \InfR-1B-Instruct outperforms Llama-3.2-1B-Instruct in these reasoning benchmarks, and is proximate to Qwen-2.5-1.5B-Instruct in mathematical reasoning and code tasks.}
\label{tab:post-trained_model_performance}
\end{table*}

\vspace{-0.1in}
\section{Experimental Results}
\vspace{-0.1in}
\subsection{Small Language Models}
\vspace{-0.1in}
\subsubsection{Benchmark and Quantitative Results}
\vspace{-0.1in}

\noindent \textbf{Benchmarks} We evaluate our model's capabilities across multiple dimensions using established benchmarks: MMLU~\cite{hendrycks2021measuringmassivemultitasklanguage} for general reasoning and knowledge spanning 57 subjects, HumanEval~\cite{peng2024humanevalxlmultilingualcodegeneration} and MBPP~\cite{austin2021programsynthesislargelanguage} for code generation and understanding in Python programming tasks, GSM8K~\cite{cobbe2021trainingverifierssolvemath} for grade school-level word problems, and MATH~\cite{hendrycks2021measuringmathematicalproblemsolving} for advanced mathematical reasoning covering algebra, geometry, and calculus.

\noindent \textbf{Pre-trained model} We evaluated the performance of SLMs using a few-shot setting. Table~\ref{tab:base_performance} compares the performance of \InfR-1B-Base with state-of-the-art open-source models. \InfR-1B-Base outperforms Llama3.2-1B and is comparable to Qwen2.5-1.5B.

\noindent \textbf{Post-trained model} We evaluated the performance of \InfR-1B-Instruct on the MMLU, GSM8K, MATH, HumanEval, and MBPP benchmarks using zero-shot prompts, as shown in Table~\ref{tab:post-trained_model_performance}. \InfR-1B-Instruct outperforms Llama-3.2-1B-Instruct on various reasoning tasks, and shows significant improvements in mathematical reasoning and coding tasks. Specifically, \InfR-1B-Instruct exceeds Llama-3.2-1B-Instruct by more than 26 points on GSM8K and 16 points on MATH, while surpassing Llama by 19 points on HumanEval and 7 points on MBPP. Notably, despite Qwen-2.5-1.5B-Instruct having more parameters than \InfR-1B-Instruct, it demonstrates comparable performance to Qwen-2.5-1.5B-Instruct in both coding and mathematical reasoning.

\subsubsection{Discussion and Insights}
\vspace{-0.1in}

\noindent During the training of the \InfR-1B, we observed several phenomena that significantly impacted the experimental results.

\noindent \textbf{Heuristic filter for reducing special patterns} During pretraining, metrics related to mathematics often showed large fluctuations. Analysis revealed that certain checkpoints generated an <eos> token with high probability when prompts ended with a colon “:”. To address this, we cleaned the mathematical web page data by removing text ending with a colon.

\noindent \textbf{Adjusting batch size according to data "width"} Initially, we trained the model with separate batches for coding and math, using smaller batch sizes for faster convergence. When combining coding, math, and general reasoning data, using the same batch size often led to gradient norm overflow and instability. Larger batch sizes help avoid erroneous gradient estimates when more domains are included, but they slow convergence. We balanced training costs by using a batch size of 2048.

\noindent \textbf{Importance of evaluation frameworks} An appropriate evaluation framework is crucial for training the base model. In early experiments, we used OpenCompass to evaluate benchmarks across all domains, but metric variations lacked consistency as training progressed. Upon reviewing the evaluation framework, we found errors in extracting the base model's completion results. We determined that EvalPlus is better for code evaluation, while Qwen2.5-Math's evaluation tool is optimal for mathematics.

\noindent \textbf{Appropriate Utilization of Synthetic Data} Synthetic data can enhance model metrics during pretraining. Initially, we incorporated synthetic data at a certain ratio to achieve higher scores in the base model. However, these higher-scoring models did not necessarily outperform those without synthetic data after the same SFT process. This may be because synthetic data often has lower perplexity, creating a distribution gap with web data, which affects model convergence. Therefore, we only introduce synthetic data during the annealing phase to prevent overtraining.

\noindent \textbf{Dependency of Data on Model Size in SFT} In SFT, we observe a strong dependency between the data and the base model. Data that performs well on large models may not achieve the same level of performance on smaller models. For smaller models, millions of data are required to achieve competitive performance.

\noindent \textbf{Reasoning Enhancement with Long CoT} OpenAI's o1~\cite{jaech2024openai} and Deepseek's R1~\cite{guo2025deepseek} significantly improve the complex reasoning capabilities across various domains such as mathematics, code, and science by extending the length of the CoT. Inspired by the test-time scaling approach, we also fine-tune \InfR-1B-Instruct on long CoT data to further enhance the model's reasoning abilities and investigate the impact of data scaling on performance. Detailed experimental results are provided in Appendix~\ref{appendix:long_cot}.

\begin{table}[htbp]
\centering
\begin{adjustbox}{width=\columnwidth}
\begin{tabular}{lccccc}
\hline
\textbf{Model} & \textbf{MMMU} & \textbf{ScreenSpot} & \textbf{Android World} \\ 
\hline
Qwen2-VL-2B & 41.1 & 9.3 & - \\
Qwen2.5-VL-3B & 53.1 & 55.5 & - \\
Showui-2B & - & 75.1 & 6.90 \\
\hline
\InfR-VL-1.6B & 38.8 & 76.3 & 9.48 \\
\hline
\end{tabular}
\end{adjustbox}
\vspace{-0.1in}
\caption{Performance of small multimodal models in reasoning tasks across different benchmarks}
\label{tab:multimodal_model_performance}
\end{table}

\subsection{Small Multimodal Language Model}
\noindent We evaluated the performance of \InfR-VL-1.6B on MMMU to test the general vision reasoning abilities, and then evaluated its reasoning capabilities on ScreenSpot and AndroidWorld. Table~\ref{tab:multimodal_model_performance} compares the performance of \InfR-VL-1.6B with state-of-the-art SMLM. \InfR-VL-1.6B exhibits comparable capabilities while maintaining a smaller model size.

\noindent During the training of the \InfR-VL-1.6B, we also have some insights which is helpful for reasoning enhanced MSLMs training.

\noindent \textbf{Curriculum  Learning} is crucial when employing pretrained ViT for multimodal training. For tasks such as grounding and OCR, which require detailed image comprehension, it is necessary to unfreeze the ViT while keeping LLM backbone frozen during training. End-to-end training may lead to early overfitting in the domain-specific dataset.

\noindent \textbf{Domain-Specific Reasoning Capabilities} are not strongly dependent on model sizes. Smaller models can achieve the required reasoning abilities if they are trained with suitable domain-specific datasets and paired with a backbone that possesses reasoning capabilities.~\cite{liu2025infiguiagent}

\vspace{-0.1in}
\section{Conclusion}
\vspace{-0.1in}
\noindent In conclusion, this paper demonstrates the potential of small language models (SLMs) and multimodal small language models (MSLMs) to provide efficient and accessible AI solutions. By developing novel training pipelines, we show that compact models can achieve competitive reasoning capabilities while significantly reducing computational costs and addressing privacy concerns. Our proposed models, \InfR-1B-Base, \InfR-1B-Instruct, and \InfR-VL-1.6B, achieve state-of-the-art performance, underscoring the feasibility of deploying these models on edge devices for real-world applications. This research paves the way for more sustainable and inclusive AI development, promoting broader adoption and innovation in the field.

\vspace{-0.1in}
\section{Limitation}
\vspace{-0.1in}

Due to constraints in computational resources and time, our experiments were primarily conducted on standard benchmarks. Our research focuses on the technical aspects of traditional information extraction tasks, without addressing social or security issues. We adhere to ethical standards by avoiding sensitive data or applications. Although the results are promising, the method's generalization ability in real-world scenarios require further exploration. This limitation presents opportunities for future research to validate and expand the method's applicability under more complex and diverse conditions.

\bibliography{acl_latex}

\FloatBarrier  
\onecolumn
\appendix
\section*{Appendix}

\section{General Reasoning Instruction Tuning}
\label{appendix:general_sft}
\subsection{Collection of high-quality datasets}
\noindent We used different datasets to perform SFT on the Llama-3.2 1B model with the same parameter settings and sampling size. Then, we compared the fine-tuned models' scores on MMLU and selected the datasets used by the model with the highest MMLU score. We experimented with Infinity-Instruct, orca-agentinstruct-1M-v1, tulu-3-sft-mixture, and WebInstructSub, and ultimately selected Infinity-Instruct and orca-agentinstruct-1M-v1.

\begin{table*}[htbp]
\caption{General SFT Datasets Evaluation }
\centering
\label{tab:general_sft_evaluation}
\begin{adjustbox}{width=\linewidth}
\begin{tabular}{lccccccc}
\hline
\textbf{Model} & \textbf{Epoch} & \textbf{GSM8K} & \textbf{MATH} & \textbf{MMLU} & \textbf{MBPP 0-shot} & \textbf{HumanEval}\\ 
\hline
Llama-3.2 1B Instruct & - & 43.37(44.4) & 15.98(30.6) & 47.18(49.3) & 36.0 & 39.63\\ 
\hline
Llama-3.2 1B +tulu-3-sft-mixture(939k) & 3.27 & 37.07 & 10.2 & 36.88 & 22.2 & 29.27\\ 
Llama-3.2 1B +Infinity-Instruct\_Gen(1.12M) & 4.11 & 31.16 & 8.1 & 39.77 & 14.2 & 18.9 \\
Llama-3.2 1B +WebInstructSub & 2.08 & 10.84 & 3.7 & 27.41 & 10.4 & 10.37\\
\hline
\end{tabular}
\end{adjustbox}
\end{table*}

\subsection{Data Synthetic Pipeline}

\noindent After collecting high-quality opensource data, we found that the model still lags behind the state-of-the-art (SOTA). Therefore, we constructed two synthetic data pipeline to achieve better performance. One for reasoning and the other for commonsense.

\noindent \textbf{Reasoning Data}
\noindent This pipeline can be divided into two main steps: seed data preparation and inference-guided rejection sampling.

\noindent Seed Data Preparation: We utilize queries related to reasoning from Infinity-Instruct as our seed data.

\noindent Rejection Sampling: After obtaining the seed queries, we incorporate "step by step" prompting to activate the reasoning capabilities of large language models, thereby generating responses with reasoning steps. For each query, we select the highest-scoring samples using a reward model. Finally, we apply certain rules (e.g., checking if the response includes reasoning steps) to further filter the samples.

\noindent \textbf{Knowledge Data}
\noindent This pipeline consists of four steps: evaluation, seed data preparation, query enhancement, and rejection sampling.

\noindent Evaluation:  We construct an offline evaluation set encompassing several domains. We observe the performance gap between our model and the state-of-the-art models on this set. Domains with significant gaps are prioritized for optimization.

\noindent Seed Data Preparation: Similarly, we use Infinity-Instruct as our seed data. Each query in Infinity-Instruct is labeled and of high quality. By applying rules, we associate key domains identified during evaluation with labels in Infinity-Instruct, thereby identifying seed data that can enhance commonsense capabilities.

\noindent Query Enhancement: For long-tail domains where seed data may be insufficient, we expand queries using large language models (LLMs) based on the seed data. During this rewriting process, we retrieve context from a corpus and include the standard response corresponding to the query in the prompting to help the LLM generate more valuable queries.

\noindent Rejection Sampling: Once queries are obtained, we employ LLMs to generate responses, as is common in synthetic data pipelines. Subsequently, we use an open-source reward model to perform rejection sampling.

\section{Mathematical Reasoning Instruction Tuning}
\label{appendix:math_sft}
\subsection{High-quality Mathematical Data}
\noindent In the SFT stage, we evaluated the quality of public datasets—including dart-math-hard~\cite{tong2024dart}, MATH-plus~\cite{yue2024mammoth2}, MetaMathQA~\cite{yu2023metamath}, orca-math-word-problems-200k~\cite{mitra2024orca}, and ScaleQuest-Math~\cite{ding2024unleashing}—through experiments on the Llama-3.2-1B baseline model. Among these, the ScaleQuest-Math dataset achieved 69.45\% on GSM8K and 42.20\% on MATH, outperforming the Llama-3.2-1B-Instruct model.

\noindent Based on Qwen2.5-Math-1.5B-Instruction model, we trained a RL model to regenerate ScaleQuest-Math answers, thereby enabling step-by-step inference. Additionally, we employ the Qwen2.5-Math-RM-72B model for rejection sampling, which evaluates the inference quality and intermediate steps of the generated query-response pairs. 

\subsection{Mathematical Data Compression}
We utilized the Llama3.3-70B-Instruct model to annotate the difficulty of the ScaleQuest-Math dataset, following the MATH benchmark. The annotations included:
\begin{itemize}
    \item \textbf{Difficulty Labels}: Very easy, easy, medium, hard, and very hard
\end{itemize}

\noindent We performed sampling experiments on the annotated ScaleQuest-Math dataset, drawing inspiration from Dart-Math~\cite{tong2024dart}'s methodology. The experiments were structured as follows:
\begin{itemize}
    \item \textbf{Group A}: Combined "very easy" and "easy" difficulty labels, comprising approximately 483,000 items
    \item \textbf{Group B}: Combined "medium", "hard", and "very hard" difficulty labels, comprising around 350,000 items
\end{itemize}

\noindent Both groups underwent experiments under identical setups. We found that on the easier GSM8K task, the performance of the two groups was similar (61.33\% for group A and 60.5\% for group B). On the more challenging MATH task, the data from group B achieved 36.64\%, outperforming group A by 13\%. Notably, using only 35\% of the dataset volume from group B nearly matched the performance of the full dataset on both GSM8K and MATH, highlighting the effectiveness of difficulty-based data compression.

\section{Code Reasoning Instruction Tuning}
\label{appendix:code_sft}
\subsection{High-quality Code Data}
To identify high-quality code datasets, we conducted a comprehensive evaluation using several code datasets to perform Supervised Fine-Tuning (SFT) on the Llama-3.2 1B model. Specifically, we fine-tuned the model with a learning rate of  \(2 \times 10^{-5}\) and a batch size of 32. Each dataset was used to fine-tune the model for 5 epochs, and the intermediate models were saved for evaluation across various benchmarks. The detailed evaluation results are presented in Table \ref{tab:code_sft_evaluation}. The Llama-3.2 1B model fine-tuned by the combination of opc-sft-stage2 and ScaleQuest-Code shows the best overall performance on the MBPP and HumanEval, which demonstrates the relatively high quality of the code datasets.

\begin{table*}[htbp]
\caption{Code SFT Evaluation (Batchsize = 32, Full SFT, Learning Rate=\num{2e-5})}
\centering
\label{tab:code_sft_evaluation}
\begin{adjustbox}{width=\linewidth}
\begin{tabular}{lccccccc}
\hline
\textbf{Model}  & \textbf{GSM8K} & \textbf{MATH} & \textbf{MMLU} & \textbf{MBPP 0-shot} & \textbf{HumanEval}\\ 
\hline
Llama-3.2 1B Instruct  & 43.37(44.4) & 15.98(30.6) & 47.18(49.3) & 36.0 & 39.63\\ 
\hline
Llama-3.2 1B + code\_instructions\_122k (122k) & 1.9 & 1.08 & 25.31 & 5.4 & 16.46\\ 
Llama-3.2 1B + stack-exchange-instruction (150k)  & 1.36 & 1.1 & 25.06 & 2.4 & 3.66\\ 
Llama-3.2 1B + magicoder-oss (75k)  & 3.11 & 1.18 & 24.61 & 17 & 9.15\\ 
Llama-3.2 1B + ScaleQuest-Code (156k) & 4.55 & 2.74 & 27.15 & 31 & 48.17\\ 
Llama-3.2 1B + opc-sft-stage2 (436k)  & 3.34 & 0.5 & 26.46 & 40.6 & 54.88\\ 
Llama-3.2 1B + opc-sft-stage2 + magicoder-oss  & 3.34 & 1.34 & 23.92 & 39.8 & 59.15\\ 
Llama-3.2 1B + opc-sft-stage2 + ScaleQuest-Code & 4.7 & 1.02 & 26.67 & 40.8 & 60.98\\ 
Llama-3.2 1B + opc-sft-stage2 + magicoder-oss + ScaleQuest-Code & 3.41 & 2.38 & 25.89 & 41.4 & 57.93\\ 
\hline
\end{tabular}
\end{adjustbox}
\end{table*}

\subsection{Synthetic Code Data}

To enhance the diversity and quality of the post-training dataset, we implement a pipeline for synthesizing code-related instruction data from validated sources. We leverage a curated subset of the Opencoder annealing dataset, comprising high-quality code snippets previously validated during the annealing stage. The seed dataset consists of multilingual code snippets with English comments.
The code data are then synthesized using the Qwen2.5-Coder-Instruct-32B model, with prompts designed to generate task descriptions, analysis steps, code solutions, and the corresponding test cases based on these seed samples. To ensure the quality of generated data, we perform execution-based code verification within a sandbox environment, using a Python interpreter as our verifier. Specifically, we include self-contained code snippets that pass all unit tests in our Supervised Fine-Tuning (SFT) dataset. We also filter out synthetic code that encounters runtime timeouts, as these solutions are often inefficient and do not meet our quality standards.

\begin{figure}[!t]
  \centering
  \scriptsize
  \begin{subfigure}[t]{.48\columnwidth}
    \includegraphics[width=7cm, height=6cm]{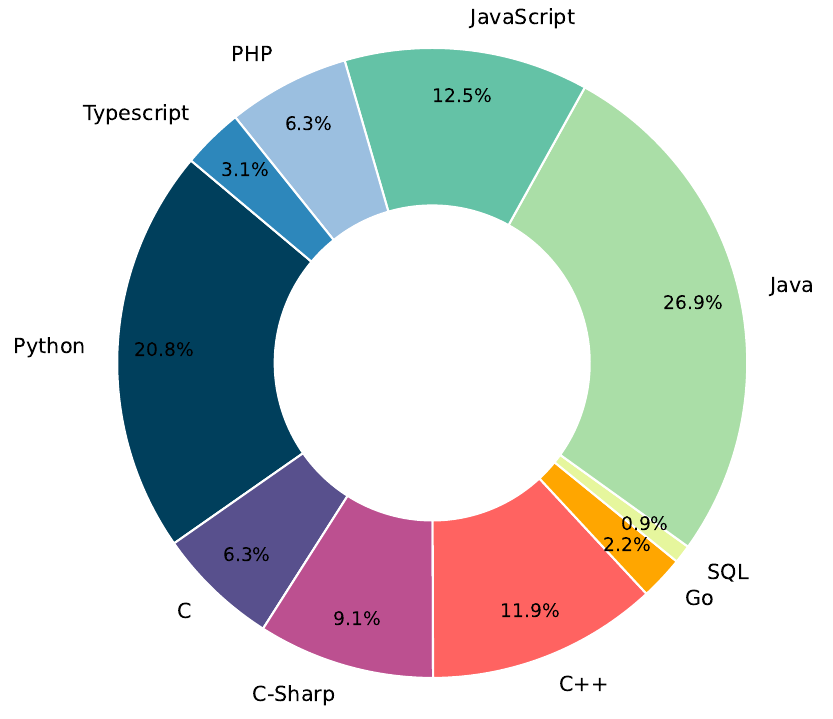}
    \caption{}
    \label{fig:pl_dis}
  \end{subfigure}
  \begin{subfigure}[t]{.48\columnwidth}
    \includegraphics[width=7cm, height=6cm]{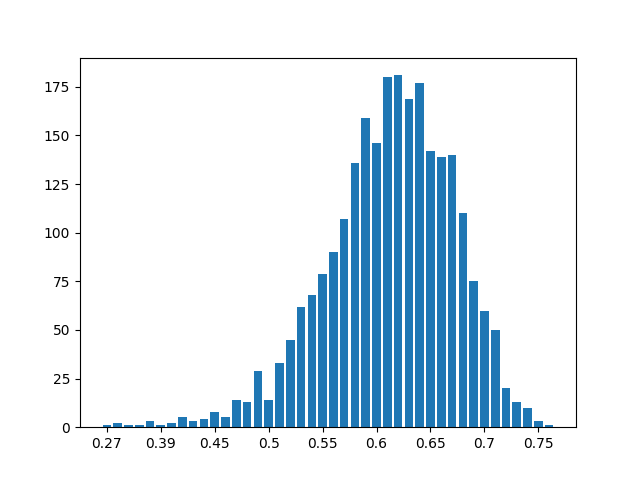}
    \caption{}
    \label{fig:mm_data_sims_ppl}
  \end{subfigure}
  \caption{Left: multi-programming language distribution. Right: similarity histogram of 2500 image-text pairs sampled from the coco-caption dataset}
\end{figure}

\begin{table*}[htpb]
\begin{adjustbox}{width=\linewidth}
\begin{tabular}{ccccccc}
\hline
\textbf{Model} & \textbf{AIME24} & \textbf{MATH500} & \textbf{AMC23} & \textbf{GPQA} & \textbf{OlympiadBench} \\ 
\hline
Llama-3.2-1B-Instruct & 0.00 & 0.25 & 0.175 & 0.02 & 0.043 \\
Qwen2.5-1.5B-Instruct & 0.067 & 0.492 & 0.225 & 0.242 & 0.185 \\
DeepSeek-R1-Distill-Qwen-1.5B & 0.289 & 0.839 & 0.700 & 0.338 & 0.436 \\
\hline
\InfR-1B-Instruct (200k Long CoT) & 0.033 & 0.474 & 0.225 & 0.288 & 0.181 \\
\InfR-1B-Instruct (2M Long CoT) & 0.067 & 0.62 & 0.300 & 0.364 & 0.224 \\
\hline
\end{tabular}
\end{adjustbox}
\caption{Results of additional fine-tuning of \InfR-1B-Instruct on Long CoT data. Further fine-tuning on Long CoT data enhances the model's complex reasoning capabilities, with performance improving as training data increases.}
\label{tab:long-cot-results}
\end{table*}

\section{Reasoning Enhancement with Long CoT}
\label{appendix:long_cot}
\noindent We selected 2M high-quality long CoT data from NuminaMath-QwQ-CoT-5M~\footnote{https://huggingface.co/datasets/PrimeIntellect/NuminaMath-QwQ-CoT-5M} to fine-tune \InfR-1B-Instruct model, further enhancing its complex reasoning performance. Referring to s1~\cite{muennighoff2025s1}'s difficulty and diversity-based sampling method, we extract a 200K subset from the full 2M dataset to assess the model's performance on data scaling. Table~\ref{tab:long-cot-results} shows the performance of the long CoT fine-tuned model on AIME24, MATH500~\cite{hendrycks2021measuringmathematicalproblemsolving}, AMC23, GPQA~\cite{rein2023gpqa} and OlympiadBench~\cite{he2024olympiadbench}. After fine-tuning on the Long CoT dataset, the model demonstrates further improvements in complex reasoning tasks. Additionally, we observe a consistent performance improvement as the training data scales from 200K to 2M, highlighting the critical role of long CoT data scale in fine-tuning small language models.


\section{Multimodal Data Cleaning}
\label{appendix:mulimodal_data_cleaning}
In order to select a suitable similarity cutoff threshold, we sampled 2,500 image-text pairs from coco-caption and calculated the similarity, and performed statistical analysis. The results are shown in Fig \ref{fig:mm_data_sims_ppl}. We can find that less than 5\% of the image-text pairs have a similarity lower than 0.5. In order to reduce the impact of these low-quality, low-similarity image-text pairs on the model's ability to understand the semantics of images and texts, we set the similarity cutoff threshold to 0.5.

\section{Data Composition}

\begin{table*}[htbp]
\centering
\begin{tabular}{ll}
    \toprule
    Type Source & Datasets \\
    \midrule
    Web Pages & Common crawl, FineWeb-Edu, DCLM \\
    Math & InfiMM-WebMath-40B, OpenWebMath, MathCode-Pile, Proof-pile-2, finemath \\
    Code & the-Stack-v2, Starcoder \\
    General Knowledge & arXiv, StackExchange, peS2o \\
    Encyclopedia & Wikipedia \\
    Open-Source Instruction & FLAN, Infinity Instruct \\
    \bottomrule
\end{tabular}
\caption{List of Pretraining data composition}
\label{tab:pretraining_data_composition}
\end{table*}

\begin{table*}[htbp]
\centering
\begin{tabular}{ll}
    \toprule
    Type Source & Datasets \\
    \midrule
    Math & InfiMM-WebMath-40B, OpenWebMath, MathCode-Pile, Proof-pile-2, finemath \\
    Code & the-Stack-v2, Starcoder, opc-annealing-corpus \\
    General Knowledge & StackExchange, peS2o\\
    Encyclopedia & Wikipedia \\
    Open-Source corpus & FLAN, Infinity Instruct, Dolmino \\
    Synthetic data & Ours \\
    \bottomrule
\end{tabular}
\caption{List of Annealing data composition}
\label{tab:annealing_data_composition}
\end{table*}

\begin{table*}[ht]
\centering
\begin{adjustbox}{width=\linewidth}
\begin{tabular}{ll}
\hline
\textbf{Task}        & \textbf{Dataset} \\
\hline
\multicolumn{2}{l}{Type: Single-image Datasets} \\
\hline
Captioning    & TextCaps, ShareGPT4o, InternVL-SA-1B-Caption, NewYorkerCaptionContest, MMInstruct \\ 
General QA    & VQAv2, GQA, OKVQA, Visual7W, MMInstruct, VSR, FSC147, Objects365-YorN, Hateful-Memes \\
OCR           & OCRVQA, TextVQA, HME100K, COCO-Text, LSVT, VCR, ST-VQA, LLaVAR, CyrillicHandwriting, IAM, NAF, \\
              & TextOCR, SROIE, MTVQA \\ 
Chart         & ChartQA, MMTab, FigureQA, VisText, ArxivQA, TabMWP, MMC-Inst, DVQA, UniChart, SimChart9K, Chart2Text \\
Document      & DocVQA, DocReason25K, Sujet-Finance-QA-Vision \\
Mathematics   & GeoQA+, CLEVR-Math, Super-CLEVR, MapQA, MAVIS, Geometry3K, TallyQA, GEOS, UniGeo, GeomVerse, \\
              & CMM-Math \\
Knowledge     & A-OKVQA, ViQuAE, iNaturalist2018, MovieNet, KonIQ-10K \\
Grounding     & RefCOCO/+/g, All-Seeing-V2, V3Det, DsLMF, COCO-ReM, TolokaVQA \\
Conversation  & ALLaVA, Viet-ShareGPT4o, Cambrain-GPT4o , RLAIF-V, Laion-GPT4V, TextOCR-GPT4V, WildVision-GPT4o \\
GUI           & Screen2Words, WebSight, Widget-Caption, RICOSCA, Seeclick, ScreenQA, \\
              & AMEX, AITW, Odyssey, UIBert, AndroidControl, Mind2Web, OmniACT, WaveUI \\
Medical       & PMC-VQA, VQA-RAD, ImageCLEF, SLAKE, VQA-Med, PathVQA \\
Science       & AI2D, ScienceQA, TQA \\
\hline
\multicolumn{2}{l}{Type: Multi-image Datasets} \\
\hline
General QA    & Img-Diff, Birds-to-Words, Spot-the-Diff, MultiVQA, NLVR2, ContrastiveCaption, DreamSim, InternVL-SA-1B-Caption \\ 
Document      & MP-DocVQA, MP-Docmatix \\
\hline
\multicolumn{2}{l}{Type: Text Datasets} \\
\hline
General QA    & UltraFeedback, UltraChat, Unnatural-Instructions, NoRobots, MOSS, LIMA, SlimOrca, WizardLM-Evol-Instruct-70K, \\
              & Llama-3-Magpie-Pro, Magpie-Qwen2-Pro, KOpen-HQ-Hermes-2.5-60K, Firefly, Dolly, OpenAI-Summarize-TLDR, \\
              & Know-Saraswati-CoT, FLAN, FLANv2 \\ 
Code          & Code-Feedback, Glaive-Code-Assistant, XCoder-80K, LeetCode, Evol-Instruct-Code \\ 
Long Context  & Long-Instruction-with-Paraphrasing, LongCite, LongQLoRA, LongAlpaca \\ 
Mathematics   & GSM8K-Socratic, NuminaMath-CoT/TIR, Orca-Math, MathQA, InfinityMATH \\ 
\hline

\end{tabular}
\end{adjustbox}
\caption{List of vision datasets used for different tasks.}
\label{tab:mm_rsn_ds}

\end{table*}

\end{document}